\documentclass[10pt,twocolumn,letterpaper]{article}

\usepackage{cvpr}              

\usepackage{graphicx}
\usepackage{amsmath}
\usepackage{amssymb}
\usepackage{multirow}
\usepackage{booktabs}
\usepackage{xcolor}
\usepackage[normalem]{ulem}
\usepackage[accsupp]{axessibility}  
\usepackage[pagebackref,breaklinks,colorlinks]{hyperref}

\usepackage[capitalize]{cleveref}
\crefname{section}{Sec.}{Secs.}
\Crefname{section}{Section}{Sections}
\Crefname{table}{Table}{Tables}
\crefname{table}{Tab.}{Tabs.}

\def\cvprPaperID{3021} 
\def\confName{CVPR}
\def\confYear{2023}

\newcommand{\better}[1]{\textcolor[rgb]{0, 0.8, 0}{#1}}
\newcommand{\worse}[1]{\textcolor[rgb]{0.8, 0, 0}{#1}}
\newcommand{\second}[1]{\textcolor[rgb]{0, 0, 1.0}{#1}}
\newcommand{\TX}[1]{\textcolor{black}{#1}}

\begin{document}

\title{CXTrack: Improving 3D Point Cloud Tracking with Contextual Information}

\author{Tian-Xing Xu$^{1}$\quad Yuan-Chen Guo$^{1}$\quad Yu-Kun Lai$^{2}$ \quad Song-Hai Zhang$^{1}$ \thanks{corresponding author}\\
$^{1}$ Tsinghua University, China \quad $^{2}$ Cardiff University, United Kingdom\\
{\tt\small $^{1}$\{xutx21@mails., guoyc19@mails.,  shz@\}tsinghua.edu.cn\quad $^{2}$LaiY4@cardiff.ac.uk}
}
\maketitle

\begin{abstract}

3D single object tracking plays an essential role in many applications, such as autonomous driving. It remains a challenging problem due to the large appearance variation and the sparsity of points caused by occlusion and limited sensor capabilities. Therefore, contextual information across two consecutive frames is crucial for effective object tracking. However, points containing such useful information are often overlooked and cropped out in existing methods, leading to insufficient use of important contextual knowledge. 
To address this issue, we propose CXTrack, a novel transformer-based network for 3D object tracking, which exploits \textbf{C}onte\textbf{X}tual information to improve the tracking results. 
Specifically, we design a target-centric transformer network that directly takes point features from two consecutive frames and the previous bounding box as input to explore contextual information and implicitly propagate target cues. To achieve accurate localization for objects of all sizes, we propose a transformer-based localization head with a novel center embedding module to distinguish the target from distractors. 
Extensive experiments on three large-scale datasets, KITTI, nuScenes and Waymo Open Dataset, show that CXTrack achieves state-of-the-art tracking performance while running at 34 FPS. 

\end{abstract}


\section{Introduction}

Single Object Tracking (SOT) has been a fundamental task in computer vision for decades, aiming to keep track of a specific target across a video sequence, given only its initial status. In recent years, with the development of 3D data acquisition devices,
it has drawn increasing attention for using point clouds to solve
various vision tasks such as object detection~\cite{he2022voxel,liu2021group,mao2021voxel,misra2021end,qi2019deep} and object tracking~\cite{qi2020p2b,wang2021mlvsnet,zheng2021box,zhou2022pttr, zheng2022beyond}.
In particular, much progress has been made on point cloud-based object tracking for its huge potential in applications such as autonomous driving~\cite{yin2021center,kuang2020probabilistic}. However, it remains challenging due to the large appearance variation of the target and the sparsity of 3D point clouds caused by occlusion and limited sensor resolution. 


\begin{figure}[t]
\centering
\includegraphics[width=1.0\linewidth]{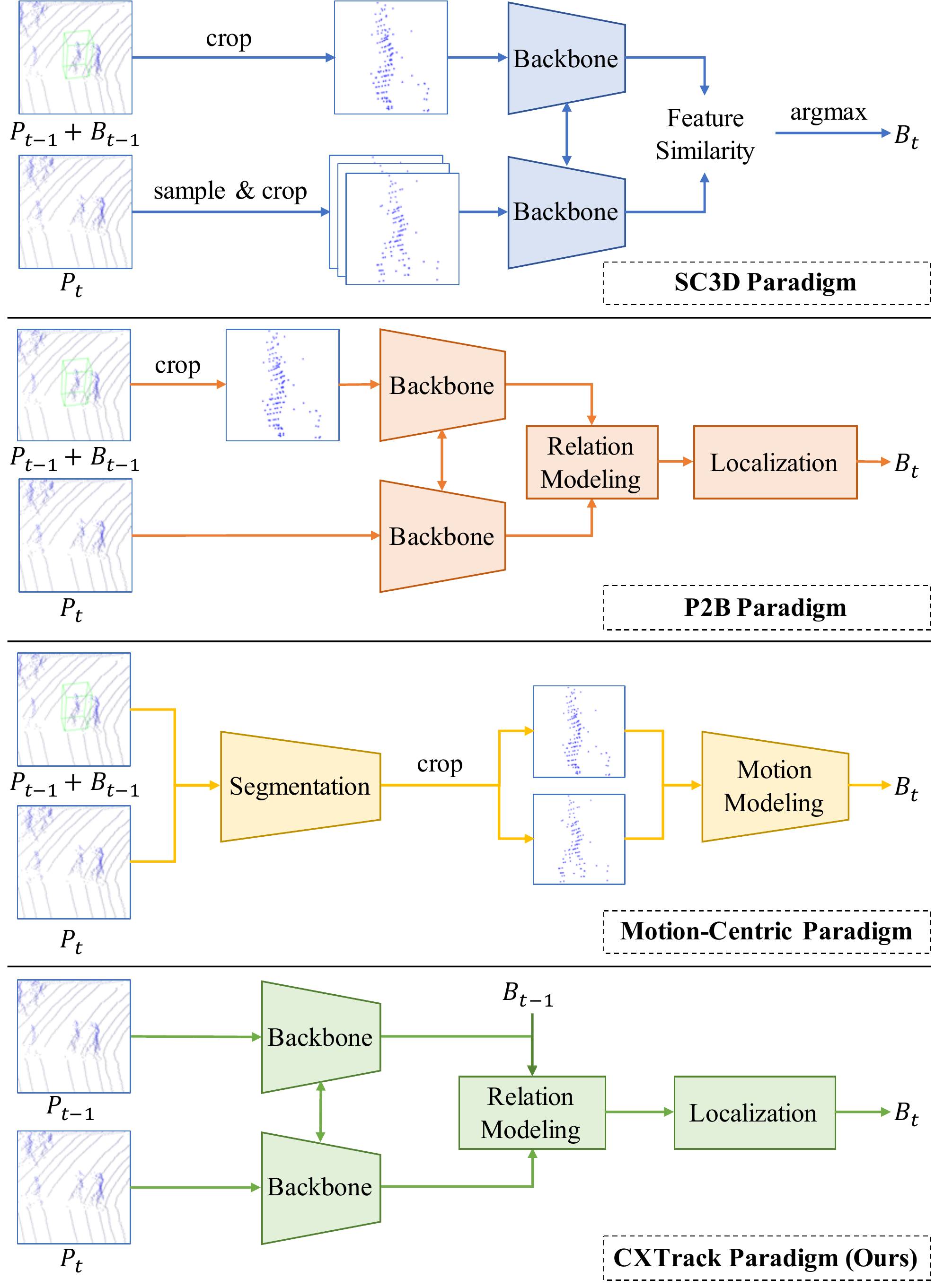}
\vspace{-0.5cm}
\caption{\textbf{Comparison of various 3D SOT paradigms.} Previous methods crop the target from the frames to specify the region of interest, which largely overlook contextual information around the target. On the contrary, our proposed CXTrack fully exploits contextual information to improve the tracking results.}
\label{fig:paradigm}
\vspace{-0.7cm}
\end{figure}

Existing 3D point cloud-based SOT methods can be categorized into three main paradigms, namely SC3D, P2B and motion-centric, as shown in \cref{fig:paradigm}. As a pioneering work, SC3D~\cite{giancola2019leveraging} crops the target from the previous frame, and compares the target template with a potentially large number of candidate patches generated from the current frame, which consumes much time. To address the efficiency problem, P2B~\cite{qi2020p2b} takes the cropped target template from the previous frame as well as the complete search area in the current frame as input, propagates target cues into the search area and then adopts a 3D region proposal network~\cite{qi2019deep} to predict the current bounding box. P2B reaches a balance between performance and speed. Therefore many follow-up works adopt the same paradigm~\cite{wang2021mlvsnet,zheng2021box,shan2021ptt,cui20213d,hui20213d,zhou2022pttr,hui20223d}. However, both SC3D and P2D paradigms overlook the contextual information across two consecutive frames and rely entirely on the appearance of the target. As mentioned in previous work~\cite{zheng2022beyond}, these methods are sensitive to appearance variation caused by occlusions and tend to drift towards intra-class distractors. To this end, M2-Track~\cite{zheng2022beyond} introduces a novel motion-centric paradigm, which directly takes point clouds from two frames without cropping as input, and then segments the target points from their surroundings. After that, these points are cropped and the current bounding box is estimated by explicitly modeling motion between the two frames. Hence, the motion-centric paradigm still works on cropped patches that lack contextual information in later localization. In short, none of these methods could fully utilize the contextual information around the target to predict the current bounding box, which may degrade tracking performance due to the existence of large appearance variation and widespread distractors.

To address the above concerns, we propose a novel transformer-based tracker named CXTrack for 3D SOT, which exploits contextual information across two consecutive frames to improve the tracking performance. As shown in \cref{fig:paradigm}, different from paradigms commonly adopted by previous methods, CXTrack directly takes point clouds from the two consecutive frames as input, specifies the target of interest with the previous bounding box and predicts the current bounding box without any cropping, largely preserving contextual information. We first embed local geometric information of the two point clouds into point features using a shared backbone network. Then we integrate the targetness information into the point features according to the previous bounding box and adopt a target-centric transformer to propagate the target cues into the current frame while exploring contextual information in the surroundings of the target. After that, the enhanced point features are fed into a novel localization head named X-RPN to obtain the final target proposals. Specifically, X-RPN adopts a local transformer~\cite{vaswani2017attention} to model point feature interactions within the target, which achieves a better balance between handling small and large objects compared with other localization heads. 
To distinguish the target from distractors, we incorporate a novel center embedding module into X-RPN, which embeds the relative target motion between two frames for explicit motion modeling.  
Extensive experiments on three popular tracking datasets demonstrate that CXTrack significantly outperforms the current state-of-the-art methods by a large margin while running at real-time (34 FPS) on a single NVIDIA RTX3090 GPU. 

In short, our contributions can be summarized as: (1) a new paradigm for the real-time 3D SOT task, which fully exploits contextual information across consecutive frames to improve the tracking accuracy; (2) CXTrack: a transformer-based tracker that employs a target-centric transformer architecture to propagate targetness information and exploit contextual information; and (3) X-RPN: a localization head that is robust to intra-class distractors and achieves a good balance between small and large targets.

\section{Related Work}

Early methods~\cite{spinello2010layered,luber2011people,pieropan2015robust} for the 3D SOT task mainly focus on RGB-D information and tend to adopt 2D Siamese networks used in 2D object tracking with additional depth maps. However, the changes in illumination and appearance may degrade the performance of these RGB-D methods. As a pioneering work in this area, SC3D~\cite{giancola2019leveraging} crops the target from the previous frame with the previous bounding box, and then computes the cosine similarity between the target template and a series of 3D target proposals sampled from the current frame using a Siamese backbone. The pipeline relies on heuristic sampling, which is very time-consuming. 

To address these issues, P2B~\cite{qi2020p2b} develops an end-to-end framework, which first employs a shared backbone to embed local geometry into point features, and then propagates target cues from the target template to the search area in the current frame. Finally, it adopts VoteNet~\cite{qi2019deep} to generate 3D proposals and selects the proposal with the highest score as the target. P2B~\cite{qi2020p2b} reaches a balance between performance and efficiency, and many works follow the same paradigm. MLVSNet~\cite{wang2021mlvsnet} aggregates information at multiple levels for more effective target localization. BAT~\cite{zheng2021box} introduces a box-aware feature fusion module to enhance the correlation learning between the target template and the search area. V2B~\cite{hui20213d} proposes a voxel-to-BEV (Bird's Eye View) target localization network, which projects the point features into a dense BEV feature map to tackle the sparsity of point clouds. Inspired by the success of transformers~\cite{vaswani2017attention}, LTTR~\cite{cui20213d} adopts a transformer-based architecture to fuse features from two branches and propagate target cues. PTT~\cite{shan2021ptt} integrates a transformer module into the P2B architecture to refine point features. PTTR~\cite{zhou2022pttr} introduces Point Relation Transformer for feature fusion and a light-weight Prediction Refinement Module for coarse-to-fine localization. ST-Net~\cite{hui20223d} develops an iterative coarse-to-fine correlation network for robust correlation learning. 

Although achieving promising results, the aforementioned methods crop the target from the previous frame using the given bounding box. This overlook of contextual information across two frames makes these methods sensitive to appearance variations caused by commonly occurred occlusions and thus the results tend to drift towards intra-class distractors, as mentioned in M2-Track~\cite{zheng2022beyond}. To this end, M2-Track introduces a motion-centric paradigm to handle the 3D SOT problem, which directly takes the point clouds from two consecutive frames as input without cropping. It first localizes the
target in the two frames by target segmentation, and then adopts PointNet~\cite{qi2017pointnet} to predict the relative target motion from the cropped target area that lacks contextual information. M2-Track could not fully utilize local geometric and contextual information for prediction, which may hinder precise bounding box regression.

\section{Method}

\subsection{Problem Definition}

Given the initial state of the target, single object tracking aims to localize the target in a dynamic scene frame by frame. The initial state in the first frame is given as the 3D bounding box of the target, which can be parameterized by its center coordinates ($x, y, z$), orientation angle $\theta$ (around the up-axis, which is sufficient for most tracked objects staying on the ground)
and sizes along each axis ($w, l, h$). Since the tracking target has little change in size across frames even for non-rigid objects, we assume constant target size and only regress the translation offset ($\Delta x, \Delta y, \Delta z$) and the rotation angle ($\Delta \theta$) between two consecutive frames to simplify the tracking task. By applying the translation and rotation to the 3D bounding box $\mathcal{B}_{t-1} \in \mathbb{R}^7$ in the previous frame, we can compute the 3D bounding box $\mathcal{B}_{t}\in \mathbb{R}^7$ to localize the target in the current frame. 

Suppose the point clouds in two consecutive frames are denoted as $\mathcal{P}_{t-1} \in \mathbb{R}^{\dot{N}_{t-1} \times 3}$ and $\mathcal{P}_t \in \mathbb{R}^{\dot{N}_{t} \times 3}$, respectively, where $\dot{N}_{t-1}$ and $\dot{N}_t$ are the numbers of points in the point clouds. We follow M2-Track~\cite{zheng2022beyond} and encode the 3D bounding box $\mathcal{B}_{t-1}$ as a targetness mask $\mathcal{\dot{M}}_{t-1} = (m^1_{t-1}, m^2_{t-1}, \cdots, m^{\dot{N}_{t-1}}_{t-1}) \in \mathbb{R}^{\dot{N}_{t-1}}$ to indicate the target position, where the mask $m_{t-1}^i$ for the $i$-th point $p_{t-1}^i$ is defined as 
\begin{equation}
m_{t-1}^i = \left\{\begin{matrix}
0 & p_{t-1}^i \mathrm{\ not\ in\ } \mathcal{B}_{t-1}\\ 
1 & p_{t-1}^i \mathrm{\ in\ } \mathcal{B}_{t-1}
\end{matrix}\right.
\end{equation}
Thus, the 3D SOT task can be formalized as learning the following mapping
\begin{equation}
\mathcal{F}(\mathcal{P}_{t-1},\dot{\mathcal{M}}_{t-1}, \mathcal{P}_{t}) \mapsto (\Delta x, \Delta y, \Delta z, \Delta \theta)
\label{eq:paradigm}
\end{equation}

\subsection{Overview of CXTrack}

\begin{figure*}
\centering
\includegraphics[width=1.0\linewidth]{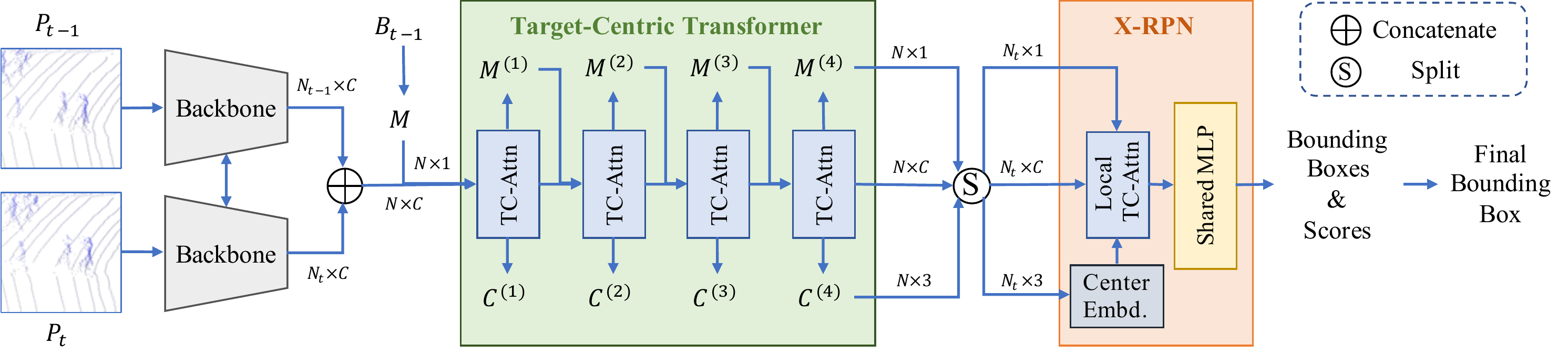}
\vspace{-0.5cm}
\caption{\textbf{The overall architecture of CXTrack.} Given two consecutive point clouds and the 3D BBox in the previous frame, CXTrack first embeds the local geometry into point features using the backbone. Then, CXTrack employs the target-centric transformer to explore contextual information across two frames and propagate the target cues to the current frame. Finally, the enhanced features are fed into a novel localization network named X-RPN to obtain high-quality proposals for verification.}
\label{fig:network}
\vspace{-0.3cm}
\end{figure*}

Following \cref{eq:paradigm}, we propose a network named CXTrack to improve tracking accuracy by fully exploiting contextual information across frames, and the overall design is illustrated in \cref{fig:network}.
We first apply a hierarchical feature learning architecture as the shared backbone to embed local geometric features of the point clouds into point features. We use $N_{t-1}$ and $N_t$ to denote the numbers of point features extracted by the backbone. 
For convenience of calculation, we create a targetness mask $\mathcal{\dot{M}}_t$ and fill it with 0.5 as it is unknown. We then concatenate the point features and targetness masks of the two frames to get $\mathcal{X} = \mathcal{X}_{t-1}\oplus \mathcal{X}_t \in \mathbb{R}^{N\times C}$ and $\mathcal{M} = \mathcal{M}_{t-1} \oplus \mathcal{M}_t \in \mathbb{R}^{N\times 1}$, where $N=N_{t-1}+N_t$, $\mathcal{M}_{t-1}$ and $\mathcal{M}_t$ are masks corresponding to point features, and extracted from $\mathcal{\dot{M}}_{t-1}$ and $\mathcal{\dot{M}}_t$, and $C$ is the number of channels for point features. We employ the target-centric transformer (\cref{sec:transformer}) to integrate the targetness mask information into point features while exploring the contextual information across frames. Finally, we propose a novel localization network, named X-RPN (\cref{sec:x-rpn}), to obtain the target proposals. The proposal with the highest targetness score is verified as the result.

\subsection{Target-Centric Transformer}

\label{sec:transformer}

\begin{figure*}
    \centering
    \begin{subfigure}{0.28\linewidth}
        \centering
        \includegraphics[width=\linewidth]{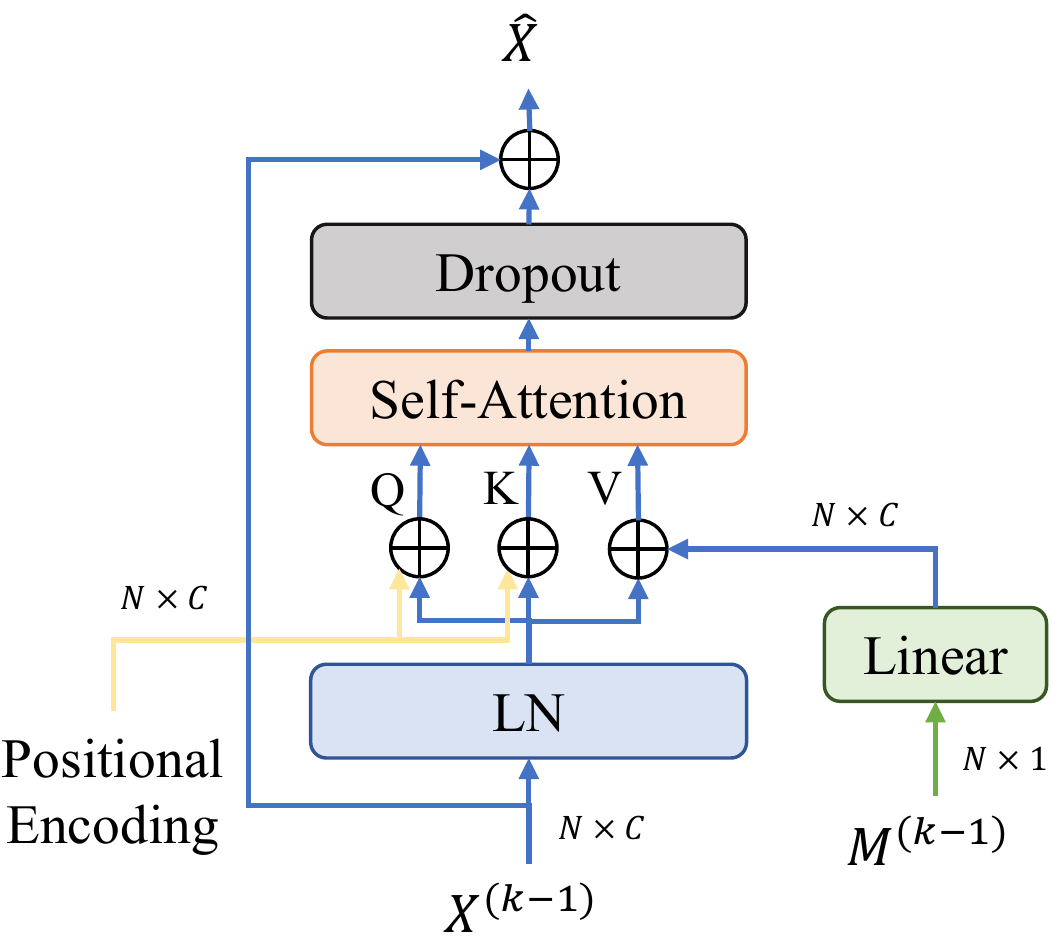}
        \caption{Vanilla}
        \label{fig:attn_vanilla}
    \end{subfigure}
    \begin{subfigure}{0.32\linewidth}
        \centering
        \includegraphics[width=\linewidth]{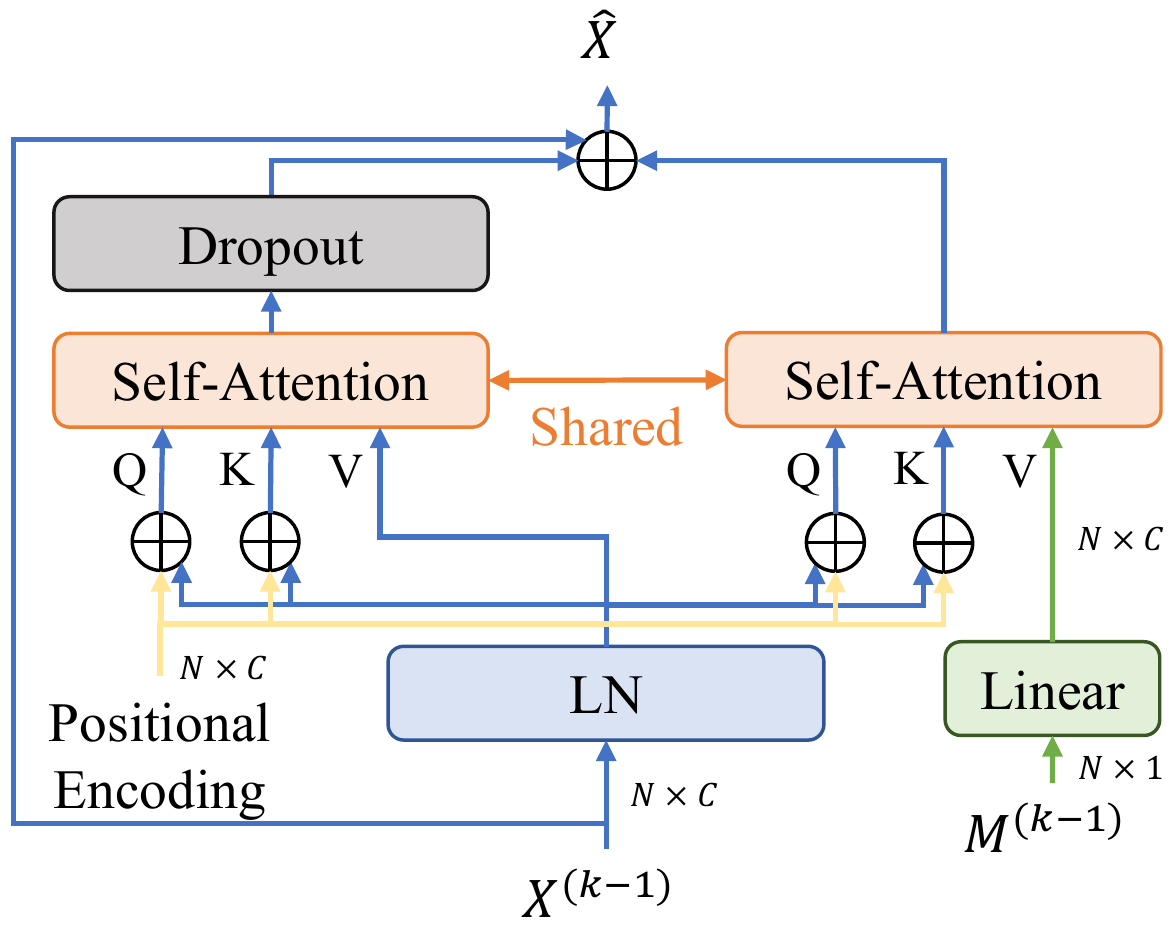}
        \caption{Semi-Dropout}
        \label{fig:attn_semi}
    \end{subfigure}
    \begin{subfigure}{0.39\linewidth}
        \centering
        \includegraphics[width=\linewidth]{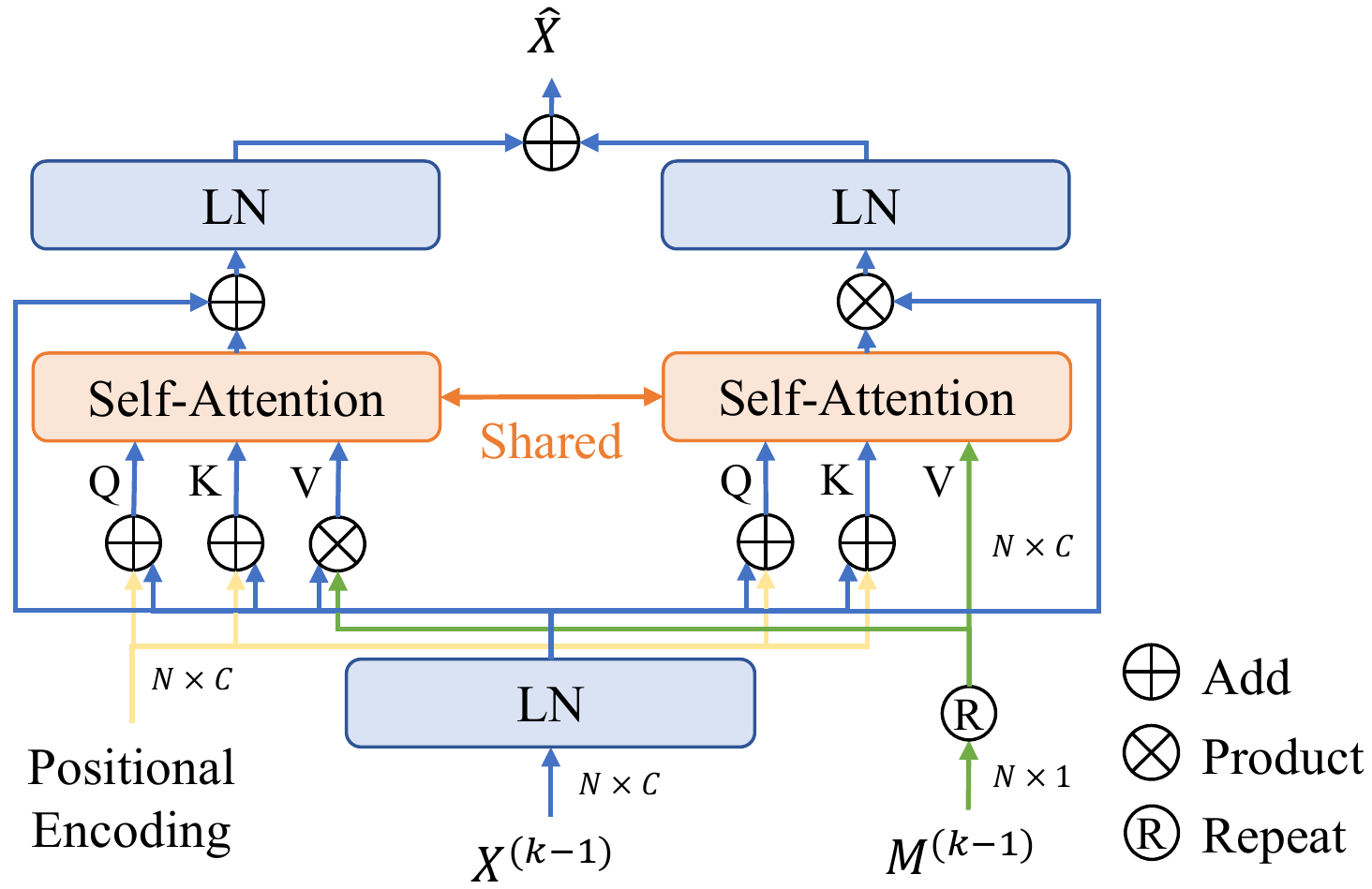}
        \caption{Gated}
        \label{fig:attn_gated}
    \end{subfigure}
    
    \vspace{-0.2cm}
    \caption{\textbf{Comparison of various transformer layers to fuse the targetness mask and point features.} We introduce three types of target-centric transformer layers, namely Vanilla, Semi-Dropout and Gated layer to integrate the targetness mask information into the point features while modeling intra-frame and inter-frame feature relationships.}
    \label{attn}
\vspace{-0.5cm}
\end{figure*}

Target-Centric Transformer aims to enhance the point features using the contextual information around the target while propagating the target cues from the previous frame to the current frame. It is composed of $N_L=4$ identical layers stacked in series.
Given the point features $\mathcal{X}^{(k-1)} \in \mathbb{R}^{N\times C}$ and the targetness mask $\mathcal{M}^{(k-1)}$ from the $(k-1)$-th layer as input ($\mathcal{M}^{(0)} = \mathcal{M}$ and $\mathcal{X}^{(0)} = \mathcal{X}$), the $k$-th layer first models the interactions between any two points while integrating the targetness mask into point features using a modified self-attention operation, and then adopts Multi-Layer-Perceptrons (MLPs) to compute the new point features $\mathcal{X}^{(k)}$ as well as the refined targetness mask $\mathcal{M}^{(k)}$. 
Thus, the predicted targetness mask will be consistently refined layer by layer. Moreover, we found it beneficial to add an auxiliary loss by predicting a potential target center for each point via Hough voting, so each layer also applies a shared MLP to generate the potential target center $C^{(k)}\in \mathbb{R}^{N\times 3}$. 

Formally, we first employ layer normalization~\cite{ba2016layer} $\text{LN}(\cdot)$ before the self-attention mechanism~\cite{vaswani2017attention} following the design of 3DETR~\cite{misra2021end}, which can be written as
\begin{equation}
    \overline{X} = \text{LN}(\mathcal{X}^{(k-1)})
\end{equation}
Then, we add positional encodings (PE) of the coordinates to the normalized point features before feeding them into the self-attention operation
\begin{align}
    X_Q = X_K &= \overline{X} + \text{PE} \\
    X_V &= \overline{X}
    \label{eq:xv}
\end{align}
It is worth noting that we only adopt PE for the query and key branches, therefore each refined point feature is constrained to focus more on local geometry instead of its associated absolute position. Subsequently, the transformer layer employs a global self-attention operation to model the relationships between point features, formulated as
\begin{align}
    \text{MHA}(X_Q, X_K, X_V) &= \text{Concat}(\text{head}_1,...\text{head}_h)W^O \\
    \text{where\;\;} \text{head}_i &= \text{Attn}(X_QW^Q_i,X_KW^K_i,\overline{X}W^V_i),\\
    \text{Attn}(Q,K,V) &= \text{softmax}(\frac{QK^T}{\sqrt{d_{k}}})V 
\end{align}
Here, MHA indicates a multi-head attention, where the attention is applied in $h$ subspaces before concatenation. The projections are implemented by parameter matrices $W^Q_i\in \mathbb{R}^{C\times d_{k}}$, $W^K_i\in \mathbb{R}^{C\times d_{k}}$, $W^V_i\in \mathbb{R}^{C\times d_{v}}$ and $W^O_i\in \mathbb{R}^{hd_v\times C}$, where $i$ indicates the $i$-th subspace. The self-attention sublayer can be written as
\begin{align}
    \widehat{X} &= \mathcal{X}^{(k-1)} + \text{Dropout}(\text{MHA}(X_Q,X_K,X_V))
    \label{eq:self-attn}
\end{align}

In addition to the self-attention sublayer, each transformer layer also contains a fully connected feed-forward network to refine the point features. The final output of the $k$-th transformer layer is given by
\begin{align}
    \mathcal{X}^{(k)} &= \widehat{X} +  \text{Dropout}(\text{FFN}(\text{LN}(\widehat{X}))), \\
\text{where}\;\; \text{FFN}(x) &= \max(0, xW_1 + b_1)W_2 + b_2.
\end{align}

To integrate the targetness mask information into point features, we need to modify the classic
transformer layer. We introduce three types of modified transformer layers in \cref{attn}, namely Vanilla, Semi-Dropout and Gated layer. 

\noindent\textbf{Vanilla.} We project the input $\mathcal{M}^{(k-1)}$ to mask embedding $\text{ME} \in \mathbb{R}^{N\times C}$ using a linear transformation. Following the design of positional encoding, we simply add $\text{ME}$ to the input token embedding $X_V$, which re-formulates \cref{eq:xv} as 
\begin{equation}
    X_V = \overline{X}+\text{ME}
\end{equation}

\noindent\textbf{Semi-Dropout.} Notably, the targetness mask information can only flow across layers along the attention path. For small objects which only have a few points to track, applying dropout to the mask embedding may discard the targetness information and lead to performance degradation. To this end, we separate the self-attention mechanism into a feature branch and a mask branch with shared attention weights, while only applying dropout to the refined point features. As shown in \cref{fig:attn_semi}, the self-attention sublayer in \cref{eq:self-attn} is re-formulated as
\begin{multline}
    \widehat{X} = \mathcal{X}^{(k-1)} + \text{Dropout}(\text{MHA}(X_Q,X_K,\overline{X})) 
    \\ + \text{MHA}(X_Q,X_K,\text{ME})
\end{multline}

\noindent\textbf{Gated.} Inspired by the design of TrDimp~\cite{wang2021transformer}, we introduce a gated mechanism into the self-attention sublayer to integrate the mask information. It has two parallel branches, namely mask transformation and feature transformation. For mask transformation, we first obtain the feature mask $\overline{M}\in \mathbb{R}^{N\times C}$ by repeating the input point-wise mask $\mathcal{M}^{(k-1)} \in \mathbb{R}^{N\times 1}$ for $C$ times. Then we can propagate the targetness cues to the current frame via adopting self-attention on the mask feature. The transformed mask serves as the gate matrix for the point features
\begin{equation}
    \widehat{X}_m = \text{LN}(\text{MHA}(X_Q, X_K, \overline{M})\otimes \overline{X}).
\end{equation}
For feature transformation, we first mask the point features to suppress feature activation in background areas, and then employ self-attention with a residual connection to model the relationships between features
\begin{equation}
    \widehat{X}_f = \text{LN}(\text{MHA}(X_Q, X_K, \overline{M}\otimes \overline{X}) + \overline{X}).
\end{equation}
As illustrated in \cref{fig:attn_gated}, we sum and normalize the output features $\widehat{X}_m$ and $\widehat{X}_m$ from the two branches. \cref{eq:self-attn} can be re-formulated as
\begin{equation}
    \widehat{X} = \text{LN}(\widehat{X}_f + \widehat{X}_m).
\end{equation}

Among the above three layers, we observe significant performance gain from using Semi-Dropout target-centric transformer layers (\cref{ablation_study}). Thus CXTrack employs Semi-Dropout layers to integrate targetness information while exploring contextual information across frames.

\subsection{X-RPN}
\label{sec:x-rpn}

\begin{figure}[t]
\centering
\includegraphics[width=0.9\linewidth]{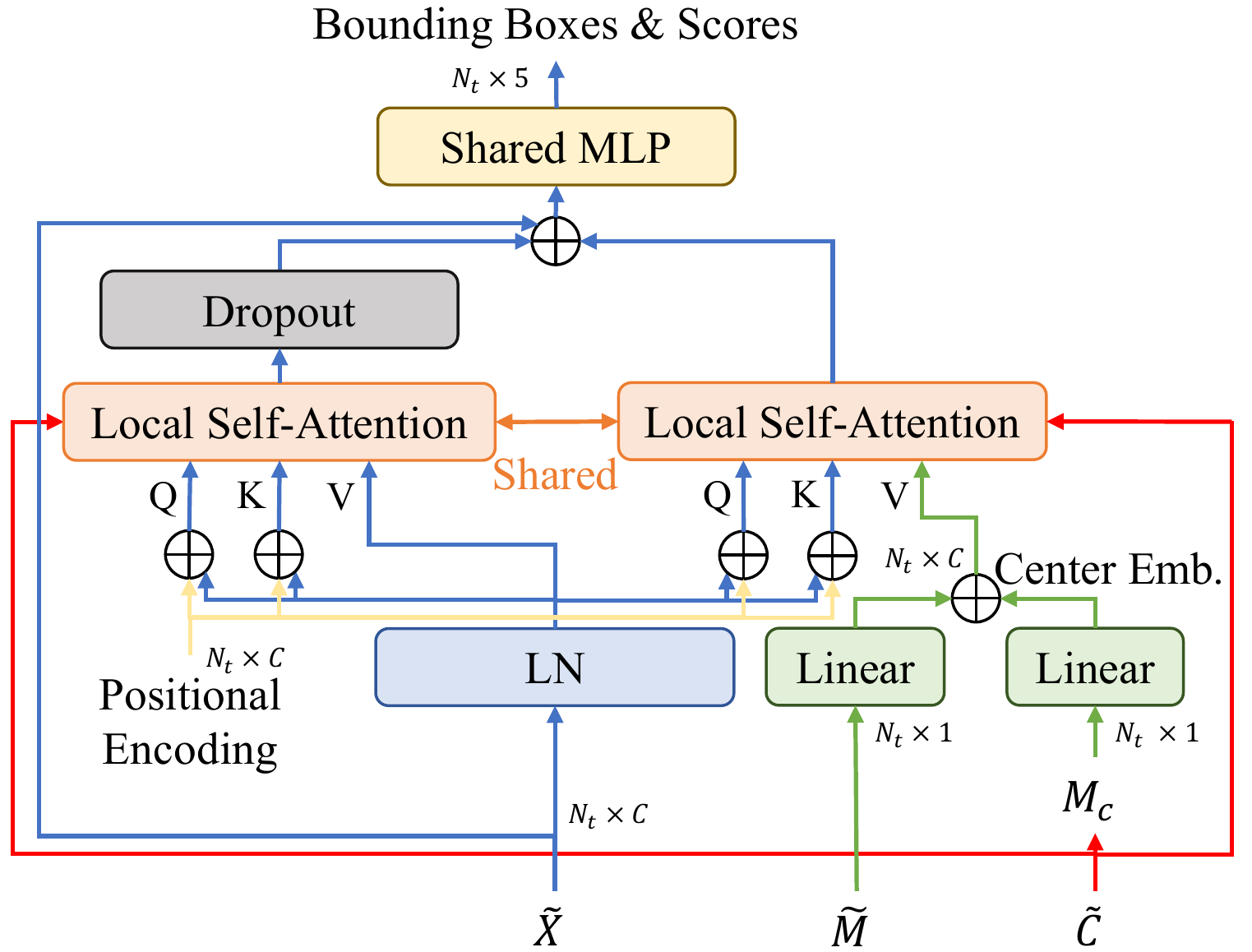}
\vspace{-0.2cm}
\caption{\textbf{The overall architecture of X-RPN.} X-RPN adopts a local transformer to model point feature interaction within the target and aggregate local clues. It also incorporates a center embedding mechanism which embeds the relative target motion between two frames to distinguish the target from distractors.}
\label{fig:xrpn}
\vspace{-0.5cm}
\end{figure}

Previous works~\cite{qi2020p2b} indicate that individual point features can only capture limited local information, which may not be sufficient for precise bounding box regression. Thus we develop a simple yet effective localization network, named X-RPN, which extends RPN~\cite{qi2019deep} using local transformer and center embedding, as shown in \cref{fig:xrpn}. Different from previous works~\cite{qi2020p2b,hui20213d}, X-RPN aggregates local clues from point features without downsampling or voxelization, thus avoiding information loss and reaching a good balance between handling large and small objects. Our intuition is that each point should only interact with points belonging to the same object to suppress irrelevant information. Given the point features $\mathcal{X}^{(N_L)}$, targetness mask $\mathcal{M}^{(N_L)}$ and target center $\mathcal{C}^{(N_L)}$ output by the target centric transformer, we first split them along the spatial dimension and only feed those belonging to the current frame into X-RPN, which is denoted as $\widetilde{\mathcal{X}} \in \mathbb{R}^{N_t\times C}$,  $\widetilde{\mathcal{C}}\in \mathbb{R}^{N_t\times 3}$ and  $\widetilde{\mathcal{M}}\in\mathbb{R}^{N_t\times1}$, respectively. X-RPN first computes the neighborhood $\mathcal{N}(p_i)$ for each point $p_i$ using its potential target center $c_i$
\begin{equation}
    \mathcal{N}(p_i) = \left \{p_j \Big|\; ||c_i - c_j||_2 < r \right\}
\end{equation}
Here $r$ is a hyperparameter indicating the size of the neighborhood. Then X-RPN adopts the transformer architecture mentioned in \cref{sec:transformer} to aggregate local information, where each point only interacts with its neighborhood points to suppress noise. We remove the feed-forward network in the transformer layer because we observe that one layer is sufficient to generate high quality target proposals. 

To deal with intra-class distractors which are widespread in the scenes~\cite{zheng2022beyond} especially for pedestrian tracking, we propose to combine the potential center information with the targetness mask. Our intuition lies in two folds. First, the tracking target keeps similar local geometry across two frames. Second, if the duration between two consecutive frames is sufficiently short, the displacement of the target is small. Therefore, we construct a Gaussian proposal-wise mask $\mathcal{M}_c$ to indicate the magnitude of the displacement of each proposal. Formally, for each point $p_i$ with the predicted target center $c_i$, the mask value $m^c_i\in \mathcal{M}_c$ is
\begin{equation}
    m^c_i = \exp(-\frac{||c_i - \overline{c}||_2^2}{2\sigma^2})
\end{equation}
where $\overline{c} \in \mathbb{R}^3$ is the target center in the previous frame and $\sigma$ is a learnable or fixed scaling factor.  
We embed the target center mask $\mathcal{M}_c$ into the center embedding matrix $\text{CE}\in \mathbb{R}^{N\times C}$ using a linear transformation, and equally combine the mask embedding and the center embedding.

\subsection{Loss Functions}

For the prediction $\mathcal{M}^{(k)}$ given by the $k$-th transformer layer, we adopt a standard cross entropy loss $\mathcal{L}_\text{cm}^{(i)}$. As for the potential target centers, we observe that it is difficult to regress precise centers for non-rigid objects such as pedestrians. Hence the predicted centers $\mathcal{C}^{(k)}$ are supervised by $L_2$ loss for non-rigid objects, and by Huber loss~\cite{ren2015faster} for rigid objects. For the target center regression loss $\mathcal{L}_\text{cc}^{(i)}$, only points in the ground truth bounding box are supervised. 

Following previous works~\cite{qi2020p2b}, proposals with predicted centers near the target center ($<0.3$m) are considered as positives and those far away ($>0.6$m) are considered as negatives. Others are left unsupervised. The predicted targetness mask is supervised via standard cross-entropy loss $\mathcal{L}_\text{rm}$ and only the bounding box parameters of positive predictions are supervised by Huber (Smooth-$L_1$) loss $\mathcal{L}_\text{box}$.

The overall loss is the weighted combination of the above loss terms
\begin{equation}
    \mathcal{L} = \gamma_1 \sum_{i=1}^{N_L} \mathcal{L}_\text{cm}^{(i)} + \gamma_2 \sum_{i=1}^{N_L} \mathcal{L}_\text{cc}^{(i)} + \gamma_3 \mathcal{L}_\text{rm} + \mathcal{L}_\text{box}
\end{equation}
where $\gamma_1$, $\gamma_2$ and $\gamma_3$ are hyper-parameters. We empirically set $\gamma_1 = 0.2$, $\gamma_2 = 1.0$, $\gamma_3 = 1.5$ for rigid objects and $\gamma_1 = 0.2$, $\gamma_2 = 10.0$, $\gamma_3 = 1.0$ for non-rigid objects.

\section{Experiments}

\subsection{Settings}

\noindent\textbf{Datasets.} We compare CXTrack with previous state of the arts on three large-scale datasets: KITTI~\cite{geiger2012kitti}, nuScenes~\cite{caesar2020nuscenes} and Waymo Open Dataset (WOD)~\cite{sun2020waymo}. KITTI contains 21 training video sequences and 29 test sequences. We follow previous work~\cite{giancola2019leveraging} and split the training sequences into three parts, 0-16 for training, 17-18 for validation and 19-20 for testing. For nuScenes, we use its validation split to evaluate our model, which contains 150 scenes. For WOD, we follow LiDAR-SOT~\cite{zhang2021modelfree} to evaluate our method, dividing it into three splits according to the sparsity of point clouds.

\noindent\textbf{Implementation Details.} We adopt DGCNN~\cite{wang2019dynamic} as the backbone network to extract local geometric information. In the X-RPN, we initialize the scaling parameter $\sigma^2=10$. Notably, we empirically fix $\sigma$ as a hyper-parameter for pedestrians and cyclists, and set it as a learnable parameter for cars and vans, since they may have larger motions. More details are provided in the supplementary material.

\noindent\textbf{Evaluation Metrics.} We use Success and Precision defined in one pass evaluation~\cite{kristan2016novel} as evaluation metrics. Success denotes the Area Under Curve (AUC) for the plot showing the ratio of frames where the 
Intersection Over Union (IOU) between the predicted and ground-truth bounding boxes is larger than a threshold, ranging from 0 to 1. Precision is defined as the AUC of the plot showing the ratio of frames where the distance between their centers is within a threshold, from 0 to 2 meters.

\subsection{Comparison with State of the Arts}

\begin{table}
\begin{center}
\caption{\textbf{Comparisons with the state-of-the-art methods on KITTI dataset}. ``Mean'' is the average result weighted by frame numbers. ``\second{Blue}'' and ``\textbf{Bold}'' denote previous and current best performance, respectively. Success/Precision are used for evaluation.}
\vspace{-0.3cm}
\label{kitti}
\resizebox{\linewidth}{!}{
\setlength{\tabcolsep}{3pt}
\begin{tabular}{c|c|c|c|c|c}
\hline 
\multirow{2}{*}{Method} & Car & Pedestrian & Van & Cyclist & Mean\\
&(6424) & (6088) & (1248) & (308) & (14068) \\
\hline
 SC3D & 41.3/57.9 & 18.2/37.8 & 40.4/47.0 & 41.5/70.4 & 31.2/48.5 \\
 P2B & 56.2/72.8 & 28.7/49.6 & 40.8/48.4 & 32.1/44.7 & 42.4/60.0 \\ 
 3DSiamRPN & 58.2/76.2 & 35.2/56.2 & 45.7/52.9 & 36.2/49.0 & 46.7/64.9 \\
 LTTR & 65.0/77.1 & 33.2/56.8 & 35.8/45.6 & 66.2/89.9& 48.7/65.8 \\
 MLVSNet & 56.0/74.0 & 34.1/61.1 & 52.0/61.4 & 34.3/44.5 & 45.7/66.7 \\ 
 BAT & 60.5/77.7 & 42.1/70.1 & 52.4/67.0 & 33.7/45.4 & 51.2/72.8 \\
 PTT & 67.8/81.8 & 44.9/72.0 & 43.6/52.5 & 37.2/47.3 & 55.1/74.2 \\
 V2B & 70.5/81.3 & 48.3/73.5 & 50.1/58.0 & 40.8/49.7 & 58.4/75.2 \\
 PTTR & 65.2/77.4 & 50.9/81.6 & 52.5/61.8 & 65.1/90.5 & 57.9/78.1 \\ 
 STNet & \second{\textbf{72.1}}/\second{\textbf{84.0}} & 49.9/77.2 & \second{58.0}/70.6 & \second{73.5}/\second{93.7} & 61.3/80.1 \\
 M2-Track & 65.5/80.8 & \second{61.5}/\second{88.2} & 53.8/\second{70.7} & 73.2/93.5 & \second{62.9}/\second{83.4} \\ 
\hline
 CXTrack & 69.1/81.6 & \textbf{67.0}/\textbf{91.5} & \textbf{60.0}/\textbf{71.8} & \textbf{74.2}/\textbf{94.3} & \textbf{67.5}/\textbf{85.3} \\
Improvement & \worse{↓3.0}/\worse{↓2.4} & \better{↑5.5}/\better{↑3.3} & \better{↑2.0}/\better{↑1.1} & \better{↑0.7}/\better{↑0.6} & \better{↑4.6}/\better{↑1.9} \\
\hline
\end{tabular} }
\end{center}
\vspace{-0.5cm}
\end{table}

\begin{table}
\begin{center}
\caption{\textbf{Robustness under scenes that contain intra-class distractors on KITTI Pedestrian category.}}
\label{robustness}
\vspace{-0.3cm}
\resizebox{0.85\linewidth}{!}{
\setlength{\tabcolsep}{3pt}
\begin{tabular}{c|c|c|c}
\hline
Method & All(6088) & Distractor-Only(3917) & Improvement \\
\hline
 PTTR & 50.9/81.6 & 44.3/70.0 & \worse{↓6.6}/\worse{↓11.6} \\
 STNet & 49.9/77.2 & 35.1/58.5 & \worse{↓14.8}/\worse{↓18.7}\\ 
 M2-Track & 61.5/88.2 & 58.0/88.4 & \worse{↓3.5}/\better{↑0.2}\\
\hline
 CXTrack & \textbf{67.0}/\textbf{91.5} & \textbf{66.1}/\textbf{91.3} & \worse{↓0.9}/\worse{↓0.3}\\
\hline
\end{tabular} 
}
\end{center}
\vspace{-0.7cm}
\end{table}

\begin{table*}
\begin{center}
\caption{\textbf{Comparison with state of the arts on Waymo Open Dataset.}}
\vspace{-0.3cm}
\label{waymo}
\resizebox{0.95\linewidth}{!}{
\begin{tabular}{c|cccc|cccc|c}
\hline 
\multirow{2}{*}{Method} & \multicolumn{4}{c|}{Vehicle(185731)} & \multicolumn{4}{c|}{Pedestrian(241752)} & \multirow{2}{*}{Mean(427483)}\\
& Easy & Medium & Hard & Mean & Easy & Medium & Hard & Mean & \\
\hline
 P2B& 57.1/65.4 & 52.0/60.7 & 47.9/58.5 & 52.6/61.7 & 18.1/30.8 & 17.8/30.0 & 17.7/29.3 & 17.9/30.1 & 33.0/43.8\\ 
 BAT& 61.0/68.3 & 53.3/60.9 & 48.9/57.8 & 54.7/62.7 & 19.3/32.6 & 17.8/29.8 & 17.2/28.3 & 18.2/30.3 & 34.1/44.4\\
 V2B& 64.5/71.5 & 55.1/63.2 & 52.0/62.0 & 57.6/65.9 & 27.9/43.9 & 22.5/36.2 & 20.1/33.1 & 23.7/37.9 & 38.4/50.1\\
 STNet& \textbf{\second{65.9/72.7}} & \textbf{\second{57.5/66.0}} & \textbf{\second{54.6/64.7}} & \textbf{\second{59.7/68.0}} & \second{29.2}/\second{45.3} & \second{24.7}/\second{38.2} & \second{22.2}/\second{35.8} & \second{25.5}/\second{39.9} & \second{40.4}/\second{52.1}\\ 
\hline
 CXTrack & 63.9/71.1 & 54.2/62.7 & 52.1/63.7 & 57.1/66.1 & \textbf{35.4/55.3} & \textbf{29.7/47.9} & \textbf{26.3/44.4} & \textbf{30.7/49.4} & \textbf{42.2/56.7}\\
 Improvement & \worse{↓2.0}/\worse{↓1.6} & \worse{↓3.3}/\worse{↓3.3} & \worse{↓3.5}/\worse{↓1.0} & \worse{↓2.6}/\worse{↓1.9} & \better{↑6.2}/\better{↑10.0} & \better{↑5.0}/\better{↑9.7} & \better{↑4.1}/\better{↑8.6} &  \better{↑5.2}/\better{↑9.5} & \better{↑1.8}/\better{↑4.6} \\
\hline
\end{tabular} }
\end{center}
\vspace{-0.8cm}
\end{table*}

We make comprehensive comparisons on the KITTI dataset with previous state-of-the-art methods, including SC3D~\cite{giancola2019leveraging}, P2B~\cite{qi2020p2b}, 3DSiamRPN~\cite{fang20203d}, LTTR~\cite{cui20213d}, MLVSNet~\cite{wang2021mlvsnet}, BAT~\cite{zheng2021box}, PTT~\cite{shan2021ptt}, V2B~\cite{hui20213d}, PTTR~\cite{zhou2022pttr}, STNet~\cite{hui20223d} and M2-Track~\cite{zheng2022beyond}. As illustrated in \cref{kitti}, CXTrack surpasses previous state-of-the-art methods, with a significant improvement of average Success and Precision. Notably, our method achieves the best performance under all categories, except for the Car, \TX{where voxel-based STNet\cite{hui20223d}surpasses us by a minor margin (72.1/84.0 v.s. 69.1/81.6). Most vehicles have simple shapes and limited rotation angles, which fit well in voxels. We argue that voxelization provides a strong shape prior, thereby leading to performance gain for large objects with simple shapes.}
The lack of distractors for cars also makes our improvement over previous methods insignificant. However, our method has a significant improvement (67.0/91.5 v.s. 49.9/77.2) on the Pedestrian category. We claim that this stems from our special design to handle distractors and our better preservation for contextual information. Besides, compared with M2-Track~\cite{zheng2022beyond}, CXTrack obtains consistent performance gains on all categories especially on the Success metric, which demonstrates the importance of local geometry and contextual information. For further analysis on the impact of intra-class distractors, we manually pick out scenes that contain Pedestrian distractors from the KITTI test split and then evaluate different methods on these scenes. As shown in \cref{robustness}, both M2-Track and CXTrack are robust to distractors, while CXTrack can make more accurate predictions.

\begin{table}
\begin{center}
\caption{\textbf{Comparison with state of the arts on nuScenes.}}
\label{nuscenes}
\vspace{-0.3cm}
\resizebox{1.0\linewidth}{!}{
\setlength{\tabcolsep}{3pt}
\begin{tabular}{c|c|c|c|c|c}
\hline
\multirow{2}{*}{Method} & Car & Pedestrian & Van & Cyclist & Mean \\
 & (15578) & (8019) & (3710) & (501) & (27808) \\
\hline
 SC3D & 25.0/27.1 & 14.2/16.2 & \second{25.7}/\second{\textbf{21.9}} & 17.0/18.2 & 21.8/23.1 \\
 P2B & 27.0/29.2 & 15.9/22.0 & 21.5/16.2 & 20.0/26.4 & 22.9/25.3 \\ 
 BAT & 22.5/24.1 & 17.3/24.5 & 19.3/15.8 & 17.0/18.8 & 20.5/23.0 \\
 V2B & 31.3/35.1 & 17.3/23.4 & 21.7/16.7 & \second{\textbf{22.2}}/19.1 & 25.8/29.0 \\
 STNet & \second{\textbf{32.2}}/\second{\textbf{36.1}} & \second{19.1}/\second{27.2} & 22.3/16.8 & 21.2/\second{\textbf{29.2}} & \second{\textbf{26.9}}/\second{30.8} \\ 
\hline
 CXTrack & 29.6/33.4 & \textbf{20.4}/\textbf{32.9} & \textbf{27.6}/20.8 & 18.5/26.8 &  26.5/\textbf{31.5} \\
 Improvement & \worse{↓2.6}/\worse{↓2.7} & \better{↑1.3}/\better{↑5.7} & \better{↑1.9}/\worse{↓1.1} & \worse{↓3.7}/\worse{↓2.4} & \worse{↓0.4}/\better{↑0.7} \\
\hline
\end{tabular} }
\end{center}
\vspace{-0.5cm}
\end{table}

To verify the genaralization ability of our method, we follow previous methods~\cite{hui20213d,hui20223d} and test the KITTI pre-trained model on nuScenes and WOD. The comparison results on WOD are shown in \cref{waymo}. It can be seen that our method outperforms others in terms of average Success and Precision with a clear margin. Notably, KITTI and WOD data are captured by 64-beam LiDARs, while nuScenes data are captured by 32-beam LiDARs. Thus it is more challenging to generalize the pretrained model on the nuScenes dataset. As shown in \cref{nuscenes}, our method achieves comparable performance on the nuScenes dataset. In short, CXTrack not only achieves a good balance between small objects and large objects, but also generalizes well to unseen scenes.

\begin{figure*}
\centering
\includegraphics[width=1.0\linewidth]{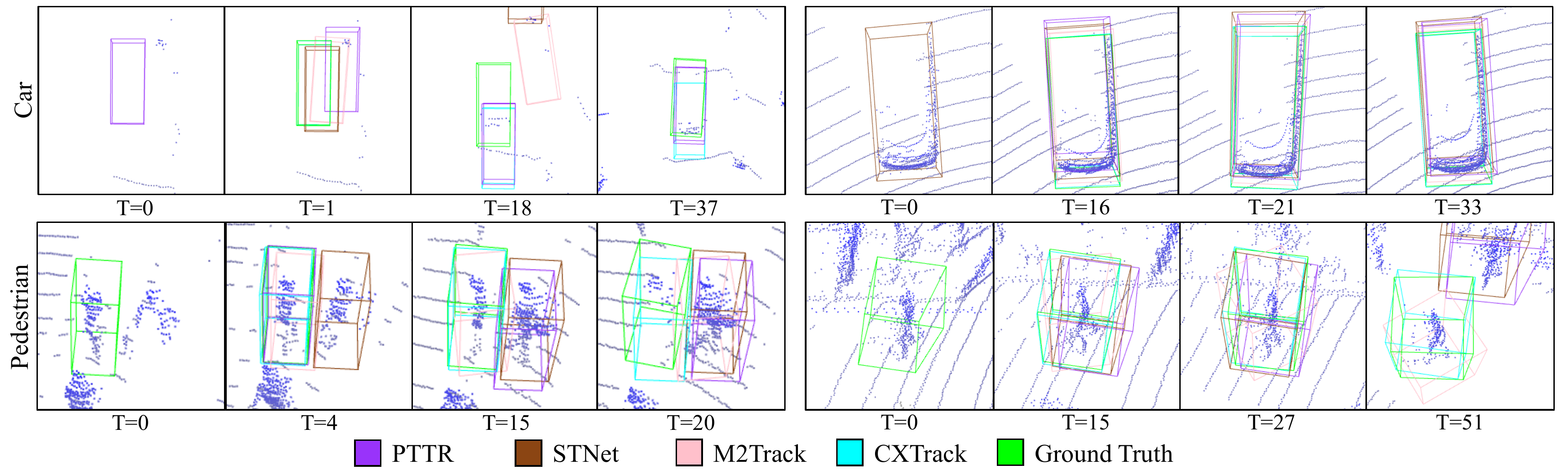}
\vspace{-0.7cm}
\caption{\textbf{Visualization results.} Left: Sparse cases in KITTI. Right: Dense cases in KITTI.}
\label{fig:visual}
\vspace{-0.5cm}
\end{figure*}

We also visualize the tracking results for qualitative comparisons. As shown in \cref{fig:visual},  CXTrack achieves good accuracy in scenes with both sparse and dense point clouds on both categories. In the sparse cases (left), previous methods drift towards intra-class distractors due to large appearance variations caused by occlusions, while only our method keeps track of the target, thanks to the sufficient use of contextual information. In the dense cases (right), our method can track the target more accurately than M2-Track by leveraging local geometric information. 

\begin{table}
\begin{center}
\caption{\textbf{Effeciency analysis of different components.} }
\label{analysis}
\vspace{-0.3cm}
\resizebox{0.8\linewidth}{!}{
\begin{tabular}{c|c|c|c}
\hline 
Component & FLOPs & \#Params & Infer Speed \\
\hline
backbone & 3.18G & 1.3M & 8.5ms\\
transformer & 1.28G & 14.7M & 10.9ms\\
X-RPN & 0.17G & 2.3M & 3.0ms\\
pre/postprocess & - & - & 6.8ms\\
 \hline
 CXTrack & 4.63G & 18.3M & 29.2ms(34FPS) \\
\hline
\end{tabular}}
\end{center}
\vspace{-0.7cm}
\end{table}

\TX{We report the efficiency of different components in \cref{analysis}. It can be observed that the target-centric transformer is the bottleneck of CXTrack during inference. We can replace the vanilla self-attention in CXTrack with linear attention such as linformer~\cite{wang2020linformer} for further speedup.}



\subsection{Ablation Studies}
\label{ablation_study}

To validate the effectiveness of several design choices in CXTrack, we conduct ablation studies on the KITTI dataset.

\begin{table}
\begin{center}
\caption{\textbf{Ablation studies of different components of the target-centric transformer.} ``Cx'' refers to contextual information, ``M'' refers to the cascaded targetness mask prediction and ``C'' refers to the auxiliary target center regression branches.}
\label{ablation_components}
\vspace{-0.2cm}
\resizebox{\linewidth}{!}{
\begin{tabular}{c|c|c|c|c|c|c|c}
\hline 
Cx & M & C & Car & Pedestrian & Van & Cyclist & Mean\\
\hline
\checkmark& & & 62.5/74.2 & 60.6/87.0 & 58.3/71.4 & 72.0/93.3 & 61.5/79.9 \\
\checkmark& \checkmark & & 67.4/80.2 & 63.9/89.0 & 57.8/70.8 & 72.7/93.8 & 65.1/83.5 \\
&\checkmark&\checkmark& 59.7/73.6${}^\dagger$ & 51.8/81.6 & 59.9/71.5 & 71.7/93.2 & 56.6/77.3\\
\checkmark&\checkmark & \checkmark & \textbf{69.1}/\textbf{81.6} & \textbf{67.0}/\textbf{91.5} & \textbf{60.0}/\textbf{71.8} & \textbf{74.2}/\textbf{94.3} & \textbf{67.5}/\textbf{85.3}\\
\hline
\end{tabular} }
\footnotesize{$\dagger$: unstable training process} 
\end{center}
\vspace{-0.6cm}
\end{table}
\noindent\TX{\textbf{Components of Target-Centric Transformer.}
\cref{ablation_components} presents ablation studies of
different components of transformer to gain a better understanding of its designs. We crop the input point cloud $\mathcal{P}_{t-1}$ using $\mathcal{B}_{t-1}$ to ablate contextual information in the previous frame. We can observe significant performance drop when not using contextual information, especially on Car and Pedestrian. For Car, it suffers from heavy occlusions(\cref{fig:visual}), while pedestrian distractors are widespread in the scene. We also find that removing context leads to unstable training on Car. We presume that the lack of supervised signals to tell the model what not to attend may confuse the model and introduce noise in training. For the cascaded targetness mask prediction and auxiliary target center regression, removing any of them leads to a obvious decline on terms of average metrics. We argue that the auxiliary regression loss can increase the feature similarities of points belonging to the same object.
}

\begin{table}
\begin{center}
\caption{\textbf{Ablation studies of different transformer layers on KITTI.} ``V'' refers to the vanilla transformer layer and ``G'' refers to the gated transformer layer. ``S'' represents the semi-dropout transformer layer which is adopted in our proposed CXTrack.}
\label{ablation_trfm}
\vspace{-0.2cm}
\resizebox{\linewidth}{!}{
\begin{tabular}{c|c|c|c|c|c}
\hline 
 & Car & Pedestrian & Van & Cyclist & Mean\\
\hline
 V & 68.8/80.4 & 62.9/87.8 & 57.2/69.6 & 72.7/94.2 & 65.3/82.9 \\
 G & 64.8/76.9 & 64.7/91.1 & 56.2/70.5 & 70.6/93.4 & 64.1/82.8 \\
 S & \textbf{69.1}/\textbf{81.6} & \textbf{67.0}/\textbf{91.5} & \textbf{60.0}/\textbf{71.8} & \textbf{74.2}/\textbf{94.3} & \textbf{67.5}/\textbf{85.3} \\
\hline
\end{tabular} }
\end{center}
\vspace{-0.5cm}
\end{table}

\noindent\textbf{Target-Centric Transformer Layer.} \cref{ablation_trfm} shows the impact of different target-centric transformer layers. Semi-Dropout achieves better performance than Vanilla, especially on Pedestrian. Small objects often consist of fewer points, hence applying dropout directly on the targetness information in training may confuse the network and lead to sub-optimal results. Gated relies entirely on the predicted targetness mask to modulate the amount of exposure for input features, which may suffer from information loss when the targetness mask is not accurate enough.

\begin{table}
\begin{center}
\caption{\textbf{Ablation studies of various localization heads on KITTI.} ``X-RPN\textbackslash C'' indicates our proposed localization head X-RPN without center embedding.}
\vspace{-0.3cm}
\label{ablation_loc_head}
\resizebox{\linewidth}{!}{
\begin{tabular}{c|c|c|c|c|c}
\hline 
Head & Car & Pedestrian & Van & Cyclist & Mean\\
\hline
 PRM~\cite{zhou2022pttr} & 66.5/77.4 & 62.2/86.8 & 52.9/64.9 & 72.5/93.8 & 63.6/80.7  \\
 RPN~\cite{qi2019deep} & 64.1/76.9 & 59.8/88.3 & 55.0/65.6 & 68.2/92.4 & 61.5/81.2 \\
 V2B~\cite{hui20213d} & \textbf{70.5}/\textbf{82.6} & 60.1/86.7 & 58.0/69.8 & 70.5/93.3 & 64.9/83.5 \\
 X-RPN\textbackslash C & 67.8/80.3 & 65.5/89.5 & 59.9/\textbf{72.1} & 72.6/94.1 & 66.2/83.9 \\
 X-RPN & 69.1/81.6 & \textbf{67.0}/\textbf{91.5} & \textbf{60.0}/71.8 & \textbf{74.2}/\textbf{94.3} & \textbf{67.5}/\textbf{85.3}\\
\hline
\end{tabular} }
\end{center}
\vspace{-0.5cm}
\end{table}

\begin{figure}[t]
\centering
\includegraphics[width=1.0\linewidth]{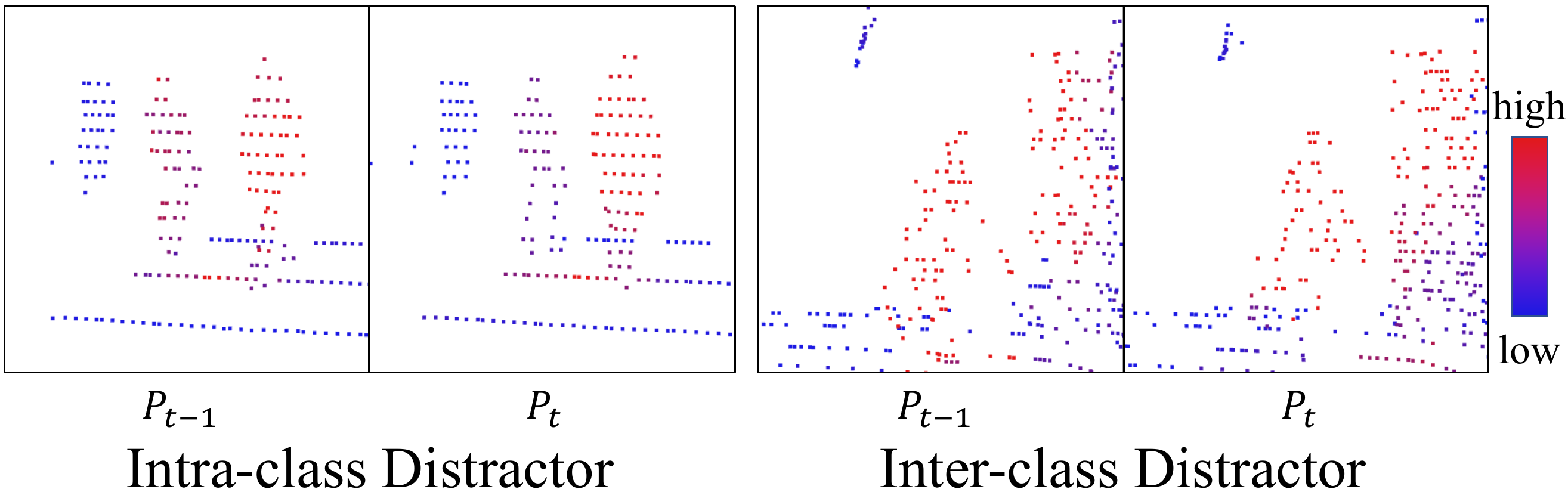}
\vspace{-0.7cm}
\caption{\textbf{Representative examples of attention maps in the transformer.} Target-centric transformer attends to objects that have similar geometry.}
\label{fig:visual_attn}
\vspace{-0.2cm}
\end{figure}

\begin{figure}[t]
\centering
\includegraphics[width=1.0\linewidth]{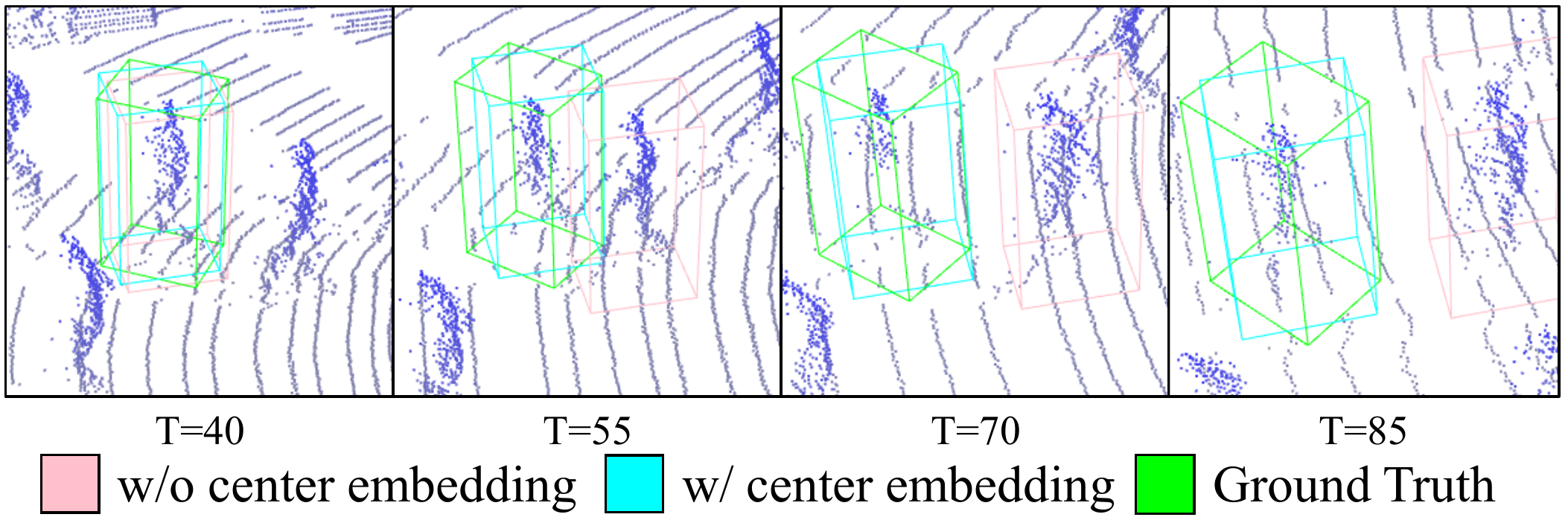}
\vspace{-0.7cm}
\caption{\textbf{Visualization of ablation study.} Center embedding can benefit object tracking in challenging scenes with distractors.}
\label{fig:visual_ce}
\vspace{-0.5cm}
\end{figure}

\noindent\textbf{X-RPN.} We replace X-RPN with other alternatives~\cite{zhou2022pttr,qi2020p2b,hui20213d} and report the comparison results in \cref{ablation_loc_head}. Although the V2B head achieves better performance than X-RPN on the Car category, it fails to track small objects such as pedestrians effectively due to intra-class distractors and inevitable information loss brought in by voxelization. It is also worth noting that we observe a performance drop without center embedding, especially on the Pedestrian category, for which distractors are more commonly seen. To explore the effectiveness of the center embedding, we visualize the attention map of the last transformer layer in \cref{fig:visual_attn}. We observe that the transformer alone can attend to regions with similar geometry to the target, but fails to distinguish the target from distractors. As shown in \cref{fig:visual_ce}, with the help of the center embedding, the network precisely keeps track of the target. In short, X-RPN achieves a good balance between large and small objects, and effectively alleviates the distractor problem.

\subsection{Failure Cases}

Although CXTrack is robust to intra-class distractors, it fails to predict accurate orientation of the target when the point clouds are too sparse to capture informative local geometry or when large appearance variations occur, as shown in \cref{fig:visual_ce}. Besides, the center embedding directly encodes the displacement of target center into features, so our model may suffer from performance degradation if trained with 2Hz data and tested with 10Hz data because the scale of the displacement differs significantly.

\section{Conclusion}

We revisit existing paradigms for the 3D SOT task and propose a new paradigm to fully exploit contextual information across frames, which is largely overlooked by previous methods. Following this paradigm, we design a novel tranformer-based network named CXTrack, which employs a target-centric transformer to explore contextual information and model intra-frame and inter-frame feature relationships. We also introduce a localization head named X-RPN to obtain high-quality proposals for objects of all sizes, as well as a center embedding module to distinguish the target from distractors. Extensive experiments show that CXTrack significantly outperforms previous state-of-the-arts, and is robust to distractors. We hope our work can promote further exploitations in this task by showing the necessity to explore contextual information for more robust predictions.

\noindent\textbf{Acknowledgment} The authors thank Jiahui Huang for his discussions. This work was supported by the Natural Science Foundation of China (Project Number 61832016), and Tsinghua-Tencent Joint Laboratory for Internet Innovation Technology.

{\small
\bibliographystyle{ieee_fullname}
\bibliography{egbib}
}

\end{document}